\documentclass[10pt,twocolumn,letterpaper]{article}

\usepackage[pagenumbers]{iccv} 

\definecolor{iccvblue}{rgb}{0.21,0.49,0.74}
\usepackage[pagebackref,breaklinks,colorlinks,allcolors=iccvblue]{hyperref}
\usepackage{multirow}

\title{TARA: Token-Aware LoRA for Composable Personalization in Diffusion Models}

\author{
Yuqi Peng$^{1,2}$
\qquad 
Lingtao Zheng$^{1}$
\qquad
Yufeng Yang$^{1}$
\qquad
Yi Huang$^{1}$
\qquad
\\
Mingfu Yan$^{1}$
\qquad
Jianzhuang Liu$^{1}$
\qquad
Shifeng Chen$^{1,3}$\thanks{Corresponding author: Shifeng Chen. This work was supported by the Shenzhen Science and Technology Program (JSGG20220831105002004) and the Shenzhen Key Laboratory of Computer Vision and Pattern Recognition.}
\vspace{0.5em}
\\
$^{1}$Shenzhen Institutes of Advanced Technology, Chinese Academy of Sciences \\
$^{2}$Northeastern University $^{3}$Shenzhen University of Advanced Technology\\
{\tt\small peng.yuq@northeastern.edu, } \\ {\tt\small \{lt.zheng2, yf.yang2, yi.huang, mf.yan, jz.liu, shifeng.chen\}@siat.ac.cn}
}

\begin{document}
\maketitle
\begin{abstract}
Personalized text-to-image generation aims to synthesize novel images of a specific subject or style using only a few reference images. Recent methods based on Low-Rank Adaptation (LoRA) enable efficient single-concept customization by injecting lightweight, concept-specific adapters into pre-trained diffusion models. However, combining multiple LoRA modules for multi-concept generation often leads to identity missing and visual feature leakage. In this work, we identify two key issues behind these failures: (1) token-wise interference among different LoRA modules, and (2) spatial misalignment between the attention map of a rare token and its corresponding concept-specific region. To address these issues, we propose Token-Aware LoRA (TARA), which introduces a token mask to explicitly constrain each module to focus on its associated rare token to avoid interference, and a training objective that encourages the spatial attention of a rare token to align with its concept region. Our method enables training-free multi-concept composition by directly injecting multiple independently trained TARA modules at inference time. Experimental results demonstrate that TARA enables efficient multi-concept inference and effectively preserving the visual identity of each concept by avoiding mutual interference between LoRA modules. The code and models are available at \href{https://github.com/YuqiPeng77/TARA}{\texttt{https://github.com/YuqiPeng77/TARA}}.
\end{abstract}
\vspace{-1.5em}    
\section{Introduction}
\begin{figure}[t]%[htbp]
\centering
\includegraphics[width=0.95\columnwidth]{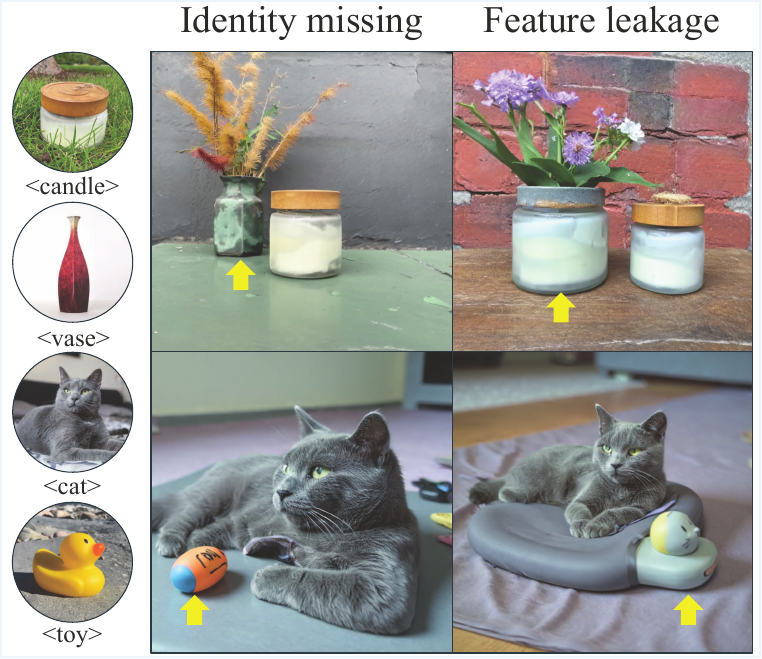} % 
\vspace{-0.5em}
\caption{Illustration of two common issues when directly applying multiple LoRA modules for multi-concept generation. With prompts ``A \(V^*_{\text{candle}}\) candle next to a \(V^*_{\text{vase}}\) vase" and ``A \(V^*_{\text{cat}}\) cat sleeps beside a \(V^*_{\text{toy}}\) toy", the left column shows \textit{identity missing}, where objects (vase and toy) are present, but their visual identities (e.g., the distinctive shape or color of \textless vase\textgreater\ and \textless toy\textgreater) are lost. The right column shows \textit{feature leakage}, where visual features of one concept (\textless candle\textgreater\ or \textless cat\textgreater) incorrectly appear on another object.}
\vspace{-1.5em}
\label{fig2}
\end{figure}
 
Diffusion models achieves remarkable progress in text-to-image generation, offering users high-quality and highly controllable image synthesis. Building on this success, a growing body of research explores how to leverage pre-trained diffusion models for personalized generation. The goal is to inject specific concepts (such as objects or styles) into the model using only 3–5 reference images, enabling the model to generate images of the personalized concept conditioned on user input (e.g., text prompt).

\begin{figure*}[t]
    \centering
    \includegraphics[width=\textwidth]{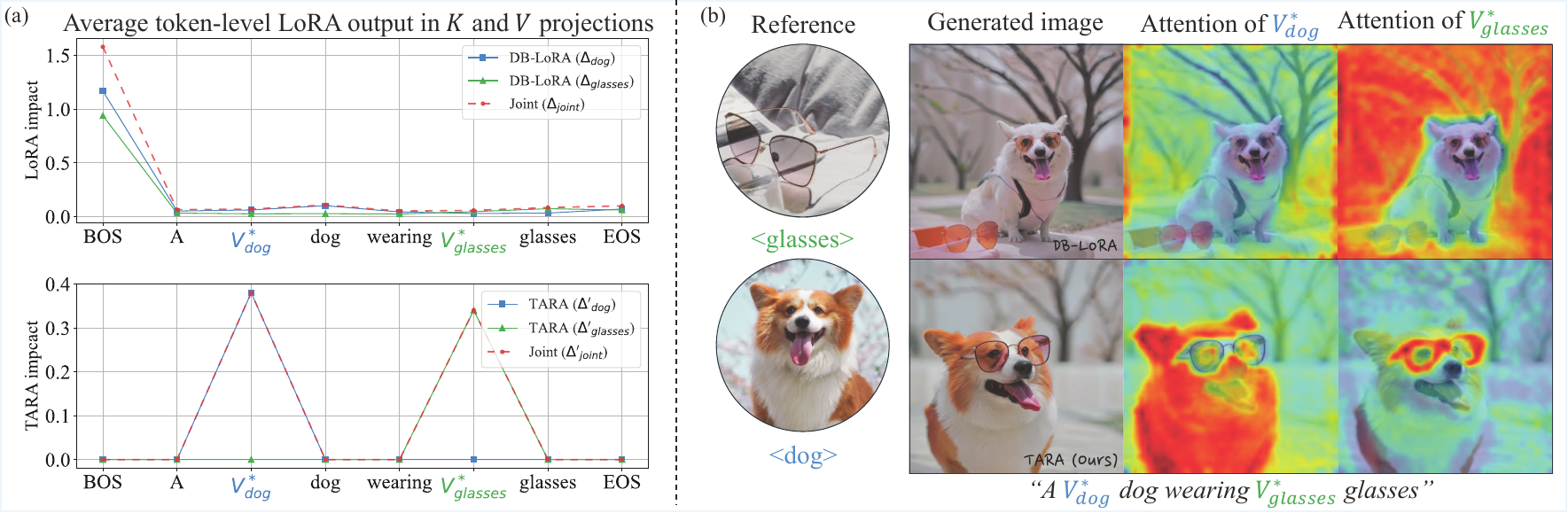}
    \vspace{-2.0em}
    \caption{Illustration of the two key issues in multi-concept generation and how TARA addresses them. (a) Visualization of the average magnitude of LoRA output in Key and Value projections of each token in the cross-attention layers. The result on top shows DreamBooth LoRA modules, which all mainly affect the \texttt{[BOS]} token, resulting in overlapping influences and potential conflicts between concepts. In contrast, TARA (bottom) uses a mask to ensure that each module focuses only on its associated rare token, effectively avoiding the interference. (b) Visualization of the spatial attention of each rare token using DAAM~\cite{tang2022daam}. DreamBooth LoRA (top) causes dispersed and overlapping attention across concepts, while TARA (bottom) tightly aligns each rare token’s attention with the correct spatial region of its corresponding concept, reducing feature entanglement.}
    \vspace{-1.5em}
    \label{fig:teaser}
\end{figure*}

To enable personalized generation, various approaches have been proposed. Methods such as DreamBooth~\cite{ruiz2023dreambooth} inject concepts by fine-tuning the entire UNet and bind each concept to a rare token, such as ``xlo", while others like Textual Inversion~\cite{gal2022image} optimize a new token embedding to represent the personalized concept. Some works~\cite{pang2024attndreambooth, kumari2023multi} further combine embedding learning with model fine-tuning. Among these approaches, Low-Rank Adaptation (LoRA)~\cite{hu2022lora} has emerged as a popular and efficient solution for personalization tasks. By inserting trainable low-rank matrices, LoRA adapts pre-trained models without modifying their original parameters. Moreover, each personalized concept can be encapsulated in a compact set of LoRA weights, significantly reducing storage overhead and enabling efficient model sharing across users.

Although LoRA achieves promising results for single-concept customization, it still faces challenges in multi-concept composition, such as identity missing and visual feature leakage. As shown in Figure~\ref{fig2}, when generating an image with two concepts, the model may generate a vase and a toy, but it cannot preserve their personalized identities and shows visual feature leakage, leading to failed multi-concept generation. While some existing works explore various fusion strategies, they often require additional training~\cite{gu2023mix,shah2024ziplora}, additional manual conditions or adapters to control the spatial placement of each concept~\cite{po2024orthogonal}, or are limited to combining style and object LoRAs only~\cite{frenkel2024implicit,ouyang2025k}. In contrast to these approaches, our goal is to develop a lightweight and practical solution that enables multi-concept generation without any additional training or external conditions. To better understand these challenges, we investigate the behavior of LoRA modules in cross-attention layers during multi-concept generation.

% By introducing simple constraints during single-concept LoRA training, we make it possible to efficiently combine multiple independently trained LoRA modules at inference time without interference, enabling multi-concept generation purely from text prompts.

Our approach is motivated by two key insights observed during multi-concept generation using LoRA, as shown in Figure~\ref{fig:teaser}. First, although each concept is associated with a rare token in the prompt, we observe that the LoRA module exerts limited influence on the rare token within the cross-attention layers. Instead, their primary impact is often concentrated on the \texttt{[BOS]} (beginning-of-sequence) token. When multiple LoRA modules are injected simultaneously, their shared focus raises the risk of interference, which can lead to concept identity distortion or even missing concepts. Second, we observe that rare tokens tend to distribute their attentions broadly across the image (especially on the background), rather than concentrating on the corresponding target objects. When multiple LoRA modules are injected together, the attention regions associated with different rare tokens significantly overlap with regions of other concepts, resulting in visual feature leakage (e.g., features of concept A incorrectly appear within concept B).

To address these issues, we propose Token-Aware LoRA (TARA), which incorporates \textit{token focus masking} (TFM) and a \textit{token alignment loss} (TAL) to constrain the LoRA module in the cross-attention layers (specifically the Key and Value matrices). TFM filters out the impact of LoRA on non-target tokens during the forward pass using a binary mask, thereby focusing the module’s effect on its corresponding rare token. Meanwhile, TAL encourages the rare token to align its key representation with that of the class token, leading to better spatial attention localization by leveraging the model's prior knowledge. Experimental results demonstrate that our method achieves efficient generation of multiple personalized concepts without relying on additional conditions or fusion-specific training, while effectively preserving each concept’s identity. This enables high-fidelity image synthesis based solely on user-provided text prompts.
\vspace{-0.5em}

\section{Related Work}
\begin{figure*}[t]
    \centering
    \includegraphics[width=0.98\textwidth]{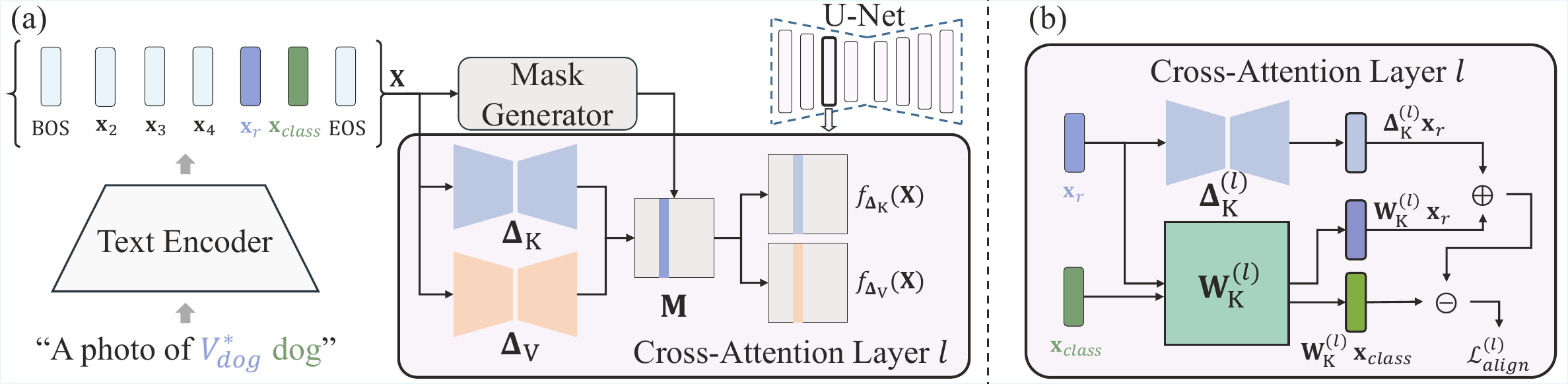} 
    \vspace{-0.5em}
    \caption{Overview of the proposed TARA framework. (a) Illustration of the Token Focus Masking (TFM). Given the prompt embeddings, a binary mask \( \mathbf{M} \) is generated to identify the rare token. The mask is applied to the outputs of the LoRA module inserted into the Key and Value paths of the cross-attention layer, producing masked outputs \( f_{\boldsymbol{\Delta}_{K}}(\mathbf{X}) \) and \( f_{\boldsymbol{\Delta}_{V}}(\mathbf{X}) \). (b) Illustration of the Token Alignment Loss (TAL). The rare token embedding \( \boldsymbol{x}_r \) is passed through both the pre-trained attention weights \( \mathbf{W}^{(l)}_{K} \) and the LoRA module \( \boldsymbol{\Delta}^{(l)}_{K} \). The sum of these outputs is then compared to the class token embedding \( \boldsymbol{x}_{\text{class}} \) processed by \( \mathbf{W}^{(l)}_{K} \). Their difference is used to compute the \( L_1 \) loss \( \mathcal{L}^{(l)}_{\text{align}} \), and the final alignment loss is the average over all layers.
}
    \vspace{-1.0em}
    \label{overview}
\end{figure*}

\paragraph{Text-to-Image Diffusion Models.}
Diffusion models are one of the foundations of modern text-to-image generation. Early methods such as DDPM~\cite{ho2020denoising} and DDIM~\cite{song2020denoising} introduce the idea of generating images by reversing a gradual noising process. Stable Diffusion~\cite{rombach2022high, podell2023sdxl} further advances this line of work by operating in a latent space, which reduces computational cost and accelerates inference. To enable text conditioning, these models are typically guided by language features extracted from pretrained encoders such as CLIP~\cite{radford2021learning} or T5~\cite{raffel2020exploring}. Thanks to its open-source availability, modular design, and strong performance, Stable Diffusion is one of the most widely used models for text-to-image generation. Today, it serves as a core backbone for many applications, including personalization~\cite{ruiz2023dreambooth, gal2022image}, image editing~\cite{hertz2022prompt, 10884879}, and controllable generation~\cite{zhang2023adding}. In this work, we also build our method on Stable Diffusion due to its broad adoption and flexibility.

\paragraph{Personalized Generation in Diffusion Models.}
Personalization aims to adapt a text-to-image model to generate images of specific subjects~\cite{shi2024instantbooth,ma2024subject}, styles~\cite{liu2024ada, wang2025sigstyle}, or identities~\cite{wu2024infinite,jang2024identity}, based on user-provided examples. Early methods~\cite{gal2022image,agarwal2025image,voynov2023p+,zhang2023prospect, ruiz2023dreambooth,fan2024dreambooth++,kumari2023multi} learn new token embeddings or fine-tune model weights to capture personalized concepts. However, these methods often require significant computational resources and are prone to overfitting or forgetting, especially when extended to multiple concepts.

Recently, LoRA-based tuning methods have been widely adopted~\cite{peng2025glad,li2024vb, chavan2023one}. Instead of updating all model parameters, LoRA injects lightweight low-rank modules into the network for efficient fine-tuning. Among them, DB-LoRA~\cite{ryu2022lora} effectively personalizes large-scale diffusion models while preserving image quality. Although these methods support multi-concept generation by merging multiple LoRA modules, they often suffer from degraded identity preservation. Mix-of-Show~\cite{gu2023mix} addresses this via additional training and external conditions, increasing complexity and limiting flexibility. Another method~\cite{po2024orthogonal} tackles the issue during single-module training by enforcing orthogonality between the basis vectors of different LoRA modules, mapping features into distinct subspaces and reducing interference. While this provides a useful perspective, it requires freezing the A matrix in each LoRA module to enforce orthogonality, reducing trainable parameters and harming generation performance.

Our method also addresses this issue from the perspective of single-concept training. Unlike prior approaches, we use Token Focus Masking (TFM) to distribute the influence of different LoRA modules across distinct tokens. The Token Alignment Loss (TAL) further concentrates each rare token’s attention on regions corresponding to its target concept, avoiding spatial conflicts. Our method enables multiple LoRA modules to be used jointly during inference without interference, effectively preserving identity consistency without additional training or input conditions.

\section{Method}

\subsection{Preliminaries}
\paragraph{Latent Diffusion Models (LDMs).}
We build our method upon the Stable Diffusion (SD V1.5) and Stable Diffusion XL (SDXL V1.0), which are latent diffusion models (LDMs) that generate high-resolution images from textual descriptions. LDMs operate by first encoding the input image into a latent space using a variational autoencoder (VAE) and then denoising a latent variable through a U-Net conditioned on text embeddings. The generation process is modeled as a Markov chain of denoising steps, aiming to learn the reverse process of a fixed forward noising schedule. The models are trained with the following objective:
\begin{equation}
\mathcal{L}_{\text{denoise}} = \mathbf{E}_{\mathbf{z_t}, \mathbf{c}, \boldsymbol{\epsilon}, t} \left[ \left\| \boldsymbol{\epsilon} - \boldsymbol{\epsilon}_{\boldsymbol{\theta}}(\mathbf{z}_t, t, \mathbf{c}) \right\|_2^2 \right],
\label{diff_obj}
\end{equation}
where \( \mathbf{z}_t \) is the noisy latent at timestep \( t \), \( \mathbf{c} \) is the text condition from a pretrained language encoder, \(\boldsymbol{\epsilon} \sim \mathcal{N}(\mathbf{0}, \mathbf{I}) \) and \( \boldsymbol{\epsilon}_{\boldsymbol{\theta}} \) is the denoising U-Net with parameters \( \boldsymbol{\theta} \).

\paragraph{DreamBooth LoRA (DB-LoRA).} DreamBooth~\cite{ruiz2023dreambooth} enhances diffusion models with personalization capabilities by injecting a new concept into the model using a few (3--5) reference images. Specifically, it associates a user-defined rare token identifier \( V^* \) (e.g., ``xlo'') with a visual concept by fine-tuning the entire U-Net of the diffusion model. Meanwhile, it leverages the prior knowledge of a class noun token \( V_{\text{class}} \) (e.g., ``dog'') corresponding to the concept's category to guide generation.

To improve training efficiency and modularity, DreamBooth can be adapted with LoRA~\cite{hu2022lora, ryu2022lora}, a parameter-efficient fine-tuning technique. LoRA inserts trainable low-rank matrices \( \boldsymbol{\Delta} = \mathbf{B}\mathbf{A} \) into all linear layers of the U-Net. During training, the modified weights become \( \mathbf{W} + \boldsymbol{\Delta} \), with \( \mathbf{A} \in \mathbb{R}^{r \times d} \), \( \mathbf{B} \in \mathbb{R}^{d \times r} \), and \( r \ll d \). Only \( \mathbf{A} \) and \( \mathbf{B} \) are updated by the same objective function in Eq.~\eqref{diff_obj}, while the original parameters \( \mathbf{W} \) remain frozen. This design significantly reduces the number of trainable parameters and enables personalized LoRA weights to be stored and transferred independently.

\begin{figure}[t]
\centering
\includegraphics[width=\columnwidth]{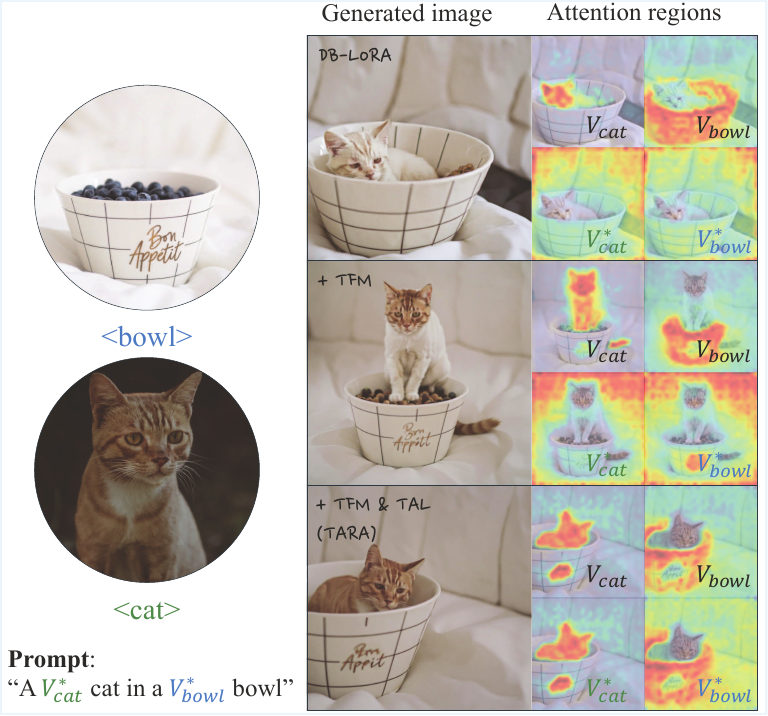} % Reduce the figure size so that it is slightly narrower than the column.
\caption{Visual examples illustrating the effect of each component in TARA. The left column shows generated images, and the right column visualizes the attention maps of class tokens (\(V_{\text{cat}}, V_{\text{bowl}}\)) and rare tokens (\(V^*_{\text{cat}}, V^*_{\text{bowl}}\)). DB-LoRA fails to reconstruct the concept identities. Applying TFM improves object identity preservation capability but feature leakage still exists. Combining TFM with TAL further enhances spatial alignment, enabling accurate reconstruction of both concept identity and visual features.}
\vspace{-1.5em}
\label{visual}
\end{figure}
\subsection{Analysis}
When multiple LoRA modules are simultaneously injected into a pre-trained diffusion model for personalized generation, the generated images often suffer from degraded quality, including \textit{identity missing} (i.e., expected concept visual identities fail to appear) and \textit{feature leakage} (i.e., visual features of different concepts undesirably blend together) as shown in Figure~\ref{fig2}. We analyze the underlying causes of these problems and identify two critical issues.

\paragraph{Identity Missing.} The core idea of DB-LoRA is to inject a LoRA module into the U-Net, binding a specific visual concept to a rare token identifier \( V^* \) and incorporating it into the model’s output domain. When a rare token appears in the prompt, the corresponding LoRA module is expected to influence the denoising trajectory, guiding the model to generate an image with the target concept. Intuitively, the injected LoRA module should focus primarily on the embedding of \( V^* \) within the cross-attention layers, so it can adjust the output of the linear layers when the rare token is present, thereby steering the denoising process. However, we observe that actually, the LoRA module does not focus on the rare token, but instead exerts considerable influence on the \texttt{[BOS]} token, as shown in Figure~\ref{fig:teaser} (a).

The incorrect token focus poses no significant problem when a single LoRA module is used for generating a single concept. This is because the LoRA module in the cross-attention layers of the U-Net relies mainly on the \texttt{[BOS]} token to guide image generation. However, when multiple LoRA modules are combined, their effects on the \texttt{[BOS]} token overlap and interfere, leading to diluted or even suppressed guidance from individual modules. As a result, the model fails to effectively steer the denoising process, ultimately causing identity missing in the output image.

Formally, let \( \mathbf{W} \) be the pre-trained Key or Value projection weight matrix of a cross-attention layer, and let \( \boldsymbol{\Delta}_i \) and \( \boldsymbol{\Delta}_j \) denote the LoRA modules trained for concepts \( i \) and \( j \), respectively. Given a text condition embedding, \(\mathbf{X} = [ \mathbf{x}_{\text{BOS}}, \mathbf{x}_2, \dots, \mathbf{x}_{r^i}, \dots, \mathbf{x}_{r^j}, \dots, \mathbf{x}_{n-1}, \mathbf{x}_{\text{EOS}} ], \) as the input to the Key or Value projection matrix in the cross-attention layer, where \( \mathbf{x}_{r^i} \) and \( \mathbf{x}_{r^j} \) are the rare tokens associated with concepts \( i \) and \( j \), respectively. When only one LoRA module \( \boldsymbol{\Delta}_i \) is injected, its main effect is concentrated on the \texttt{[BOS]} token. This effect, represented by \( \boldsymbol{\Delta}_i \mathbf{x}_{\text{BOS}} \), provides guidance for predicting noise. However, when another LoRA module \( \boldsymbol{\Delta}_j \) is injected, the effect of the \texttt{[BOS]} token becomes \(\boldsymbol{\Delta}_i \mathbf{x}_{\text{BOS}} + \boldsymbol{\Delta}_j \mathbf{x}_{\text{BOS}}\). Their effects interfere with each other, making it difficult for the model from accurately capturing the intended semantic information.

\paragraph{Visual Feature Leakage.} In the DreamBooth setup, a prompt typically follows a format that combines a rare token and a class token (i.e., \([V^*]\) \([V_{\text{class}}]\)) to represent a concept. This design is intended to help the rare token inherit the prior knowledge of the class token from the pre-trained model. However, as shown in the upper row of Figure~\ref{fig:teaser} (b), our attention analysis reveals that the spatial attention maps of rare tokens are often dispersed across the entire image. In contrast, class tokens tend to attend to more compact, concept-specific regions (see Figure~\ref{visual}). This misalignment suggests that rare tokens fail to effectively utilize the prior knowledge of their corresponding class tokens, and also fail to anchor the visual features of the concept to its correct spatial location in the generated image, potentially causing feature leakage into unrelated regions.

This issue becomes more pronounced in multi-concept generation, as the spatial attention of one rare token often overlaps with other concept-specific regions. Therefore, one concept's features may be incorrectly associated with another's region. In the upper row of Figure~\ref{fig:teaser} (b), the color associated with \textless dog\textgreater\ is polluted by the background color of \textless glasses\textgreater. We identify this phenomenon as a key contributor to visual feature leakage, where features of different subjects are mistakenly entangled.

To summarize, our analysis suggests that the primary reason why DB-LoRA modules struggle with identity preservation when injected simultaneously is twofold: different LoRA modules tend to focus on the same token \texttt{[BOS]}, and their spatial attentions often fail to localize to the correct concept regions. As a result, different modules interfere with each other by competing for influence over the same token representation and causing misalignment in their respective spatial attention regions.

\subsection{Token Focus Masking}
As discussed in the previous section, different LoRA modules tend to focus primarily on the same token (i.e., \texttt{[BOS]}) in the cross-attention layers, leading to interference. A solution is to constrain each injected LoRA module to operate solely on its corresponding rare token, preventing conflicts by localizing their effects to the distinct tokens.

To achieve this, we propose the \textit{Token Focus Masking} (TFM) strategy, which is applied to Key and Value projections in cross-attention layers during both the training of a single LoRA and the inference with multiple LoRAs, constraining the LoRA module \(\boldsymbol{\Delta_i}\) for concept \(i\) to operate solely on its associated rare token \( \mathbf{x}_{r^i} \). Specifically, we introduce a binary mask \( \mathbf{M_i} \in \{0,1\}^{d\times n} \) of the same size as the prompt embedding \( \mathbf{X} = [\mathbf{x}_1, \dots, \mathbf{x}_{r^i}, \dots, \mathbf{x}_n] \in \mathbb{R}^{d \times n} \), where \( n \) is the number of tokens and \( d \) is the embedding dimensionality. All entries of \( \mathbf{M_i} \) are set to 0 except for the column corresponding to the rare token \( \mathbf{x}_{r^i} \), whose elements are all set to 1. The mask is applied to the output of a single LoRA module during the forward pass:

\begin{equation}
f_{\boldsymbol{\Delta_i}}(\mathbf{X}) = \mathbf{M_i} \odot (\boldsymbol{\Delta_i} \mathbf{X}) = [ 0, \dots,\boldsymbol{\Delta} \mathbf{x}_{r^i}, \dots, 0 ],
\label{single_lora_mask}
\end{equation}
where \( \odot \) denotes element-wise multiplication. This ensures that only the rare token's embedding \( \mathbf{x}_{r^i} \) contributes to the LoRA output. This mechanism is shown in Figure~\ref{overview} (a).

To support multi-concept generation, we extend the token focus masking strategy to handle multiple LoRA modules simultaneously. Suppose we aim to inject \( p \) LoRA modules, \( \{ \boldsymbol{\Delta}_1, \dots, \boldsymbol{\Delta}_p \} \), each corresponding to a unique concept. For each concept \( i \in \{1, 2, \dots, p\} \), we define a rare token \( \mathbf{x}_{r^i} \) and construct a binary mask \( \mathbf{M}_i \in \{0,1\}^{d\times n} \). Let the prompt embedding be \( \mathbf{X} = [\mathbf{x}_1, \dots, \mathbf{x}_n] \in \mathbb{R}^{d \times n} \), which contains the set of rare tokens \( \{\mathbf{x}_{r^1}, \dots,  \mathbf{x}_{r^p} \}\). Each LoRA module's output is masked by its corresponding token mask, as defined in Eq.~\eqref{single_lora_mask}.

During inference, the outputs of all LoRA modules are summed and added to the output of the frozen pre-trained projection weights \( \mathbf{W} \), forming the final output:
\begin{equation}
F_{\mathbf{W},\Delta_{1,\dots,p}}(\mathbf{X}) = \mathbf{W}\mathbf{X} + \sum_{i=1}^{p} f_{\boldsymbol{\Delta}_i}(\mathbf{X}).
\label{multi_lora_mask}
\end{equation}

In summary, TFM plays a critical role in enabling conflict-free multi-concept generation. By restricting each LoRA module to operate exclusively on its corresponding rare token, the masking ensures that the functional impacts of different LoRA modules do not overlap on the same token, thereby reducing the risk of identity missing. This effect is evident in the second row of Figure~\ref{visual}, where the identities of the \textless cat\textgreater\ and \textless bowl\textgreater\ concepts are preserved.

\subsection{Token Alignment Loss}
Although the proposed TFM strategy enables the injected LoRA modules in cross-attention layers to focus on their respective rare tokens, thus mitigating interference among modules and alleviating the issue of identity missing in multi-concept generation, the problem of visual feature leakage still persists. This is because the visual feature information encoded in these tokens often fails to align with the correct spatial regions of the corresponding concepts, as shown in the second row of Figure~\ref{visual}, where the body color of \textless cat\textgreater\ is affected by \textless bowl\textgreater.

In our experiments, we observe that the concept regions in the image often align well with the attention regions of the corresponding class tokens. This suggests that if the attention region of a rare token can be aligned with that of its class token, the visual features it encodes can be more effectively anchored to the correct concept region and can help prevent feature leakage.

\begin{table*}[htbp]
\centering
\resizebox{\textwidth}{!}{
\begin{tabular}{lllccccccccc}
  \toprule
  \multirow{2}{*}[-0.1cm]{\textbf{Backbone}} & \multirow{2}{*}[-0.1cm]{\textbf{Method}} & \multirow{2}{*}[-0.1cm]{\shortstack{Merge \\ Time}} & \multicolumn{3}{c}{\textbf{CLIP-T $\uparrow$}} & \multicolumn{3}{c}{\textbf{CLIP-I $\uparrow$}} & \multicolumn{3}{c}{\textbf{DINO $\uparrow$}} \\
  \cmidrule{4-12}
  & & & Single  & Merged  & $\delta$ & Single  & Merged  & $\delta$ & Single  & Merged  & $\delta$\\
  \midrule
  \multirow{3}{*}{\shortstack{SD \\ V1.5}} 
  & Custom Diffusion & $<$1 s & .213 & .156 & -.057 & .722 & .483 & -.239 & .433 & .111 & -.322 \\
  & DB-LoRA & $<$1 s & .314 & .318 & \underline{+.004} & .743 & .678 & -.065 & .440 & .333 & -.107 \\

  & Mix-of-Show & $\sim$15 m & .278 & .303 & \textbf{+.025} & .768 & .760 & \underline{-.008} & .467 & .458 & \underline{-.009} \\
  & ROB & $<$1 s & .317 & .315 & -.002 & .712 & .702 & -.010 & .395 & .380 & -.015 \\
  
  & TARA (ours) & $<$1 s & .315 & .314 & -.001 & .780 & .776 & \textbf{-.004} & .489 & .485 & \textbf{-.004} \\
  
  \midrule
  
  \multirow{3}{*}{\shortstack{SDXL \\ V1.0}}
  & DB-LoRA & $<$1 s & .310 & .333 & \textbf{+.023} & .802 & .694 & -.108 & .535 & .356 & -.179 \\
  
  & ROB & $<$1 s & .321 & .315 & -.006 & .777 & .754 & \underline{-.023} & .495 & .462 & \underline{-.033} \\
  
  & TARA (ours) & $<$1 s & .313 & .312 & \underline{-.001} & .799 & .795 & \textbf{-.004} & .533 & .528 & \textbf{-.005} \\
  \bottomrule
\end{tabular}
}
\caption{Single-concept generation results. We provide quantitative comparisons for each method, evaluating single-concept generation performance both before and after the merging process. Our method achieves performance comparable to or better than others in the single-LoRA setting. More importantly, our method exhibits the smallest performance drop (\( \delta \)) after merging another LoRA module, indicating strong robustness against interference. All results are obtained using the same prompts.}
\vspace{-0.5em}
\label{tab:single_concept}
\end{table*}

In a cross-attention layer, each text token embedding \( \mathbf{x}_i \) is projected into a Key vector via a linear transformation: \( \mathbf{K}_i = \mathbf{W}_K \mathbf{x}_i \), forming the Key matrix \( \mathbf{K} = [\mathbf{K}_1, \dots, \mathbf{K}_n]^\top \in \mathbb{R}^{n \times d} \). The latent image features are divided into \( m \) non-overlapping patch vectors \( \{ \mathbf{f}_j \}_{j=1}^{m} \), where each patch is projected into a Query vector \( \mathbf{Q}_j = \mathbf{W}_Q \mathbf{f}_j \), forming the Query matrix \( \mathbf{Q} = [\mathbf{Q}_1, \dots, \mathbf{Q}_m]^\top \in \mathbb{R}^{m \times d} \). The spatial attention map is computed as:
\begin{equation}
\mathbf{A} = \text{softmax}\left( \frac{\mathbf{Q} \mathbf{K}^\top}{\sqrt{d}} \right),
\label{attn_map}
\end{equation}
where each column \( \mathbf{A}_i \) represents the attention distribution over spatial locations for token \( i \). Since the attention scores comes from the dot product between the Query and Key vectors, aligning the Key vector of the rare token with that of the class token helps align their spatial attention regions.

To this end, we propose the \textit{Token Alignment Loss} (TAL) to guide the injected LoRA module in the Key projections. Specifically, let \(\textbf{W}_K^{(l)}\) be the pre-trained Key projection matrix and \(\boldsymbol{\Delta}_K^{(l)}\) be the corresponding injected LoRA module in the \(l\)-th cross-attention layer. Let $\mathbf{x}_{\text{class}}$ and $\mathbf{x}_r$ respectively denote the class token and the rare token associated with the personalized concept. TAL is defined as:
\begin{equation}
\mathcal{L}_{\text{align}} = \frac{1}{L} \sum_{l=1}^{L} \left\| \mathbf{W}_K^{(l)} \mathbf{x}_{\text{class}} - (\mathbf{W}_K^{(l)} + \boldsymbol{\Delta}_K^{(l)}) \mathbf{x}_r \right\|,
\label{align_loss}
\end{equation}
where \( L \) is the total number of the cross-attention layers in the U-Net. The computation process at each layer is illustrated in Figure~\ref{overview}~(b). We combine \(\mathcal{L}_{\text{align}}\) with the diffusion model objective \(\mathcal{L}_{\text{denoise}}\) (defined in Equation~\eqref{diff_obj}) to form the final single-LoRA training objective:
\begin{equation}
\mathcal{L} = \mathcal{L}_{\text{denoise}} + \lambda \mathcal{L}_{\text{align}},
\label{total_loss}
\end{equation}
where \( \lambda \) is a weighting parameter that is set to \( 1 \).

TAL encourages the Key vector of the rare token to align with that of the class token, effectively guiding the spatial attention of the rare token to align with that of the class token. By doing so, the visual features carried by the rare token are more likely to be correctly associated to the concept region, thereby mitigating feature leakage during multi-concept generation, as shown in the third row of Figure~\ref{visual}.

\section{Experiments}
In this section, we present the experimental results of TARA in various personalized image generation settings. Both qualitative and quantitative evaluations demonstrate that our method outperforms the DB-LoRA baseline and other popular approaches in terms of identity preservation. In addition, we conduct an ablation study to analyze the individual contributions of each component in our framework.

\paragraph{Datasets.}
All experiments are conducted on the DreamBooth dataset~\cite{ruiz2023dreambooth}, which contains 30 diverse subjects, including 21 objects and 9 live subjects. Each subject is represented by 4–6 reference images. LoRA modules are individually fine-tuned using their corresponding image sets with a text prompt in the form of ``A \( V^* \space V_{\text{class}}. \)'' (e.g., ``A \( V^*_{\text{dog}} \) dog.'') to learn personalized visual representations.

\paragraph{Implementation Details.}
Our method is implemented on both Stable Diffusion V1.5 and Stable Diffusion XL 1.0 backbones. For each concept, we fine-tune only the U-Net using a LoRA module with rank 8, a learning rate of 1e-5, and a batch size of 1 for 1000 epochs. Unlike some of previous methods, both the text encoder and the text embeddings in TARA are kept frozen during training. The token alignment loss weight is set to 1 by default. All experiments are conducted on one single NVIDIA RTX A6000 GPU.

\paragraph{Baselines.}
We compare our method with Custom Diffusion~\cite{kumari2023multi}, DB-LoRA~\cite{ryu2022lora}, Mix-of-Show~\cite{gu2023mix}, and the Randomized Orthogonal Basis (ROB) method from prior works~\cite{po2024orthogonal,zhang2025lori}. Since the implementation of~\cite{po2024orthogonal} is unavailable and~\cite{zhang2025lori} focuses on language models, we re-implement ROB by initializing the LoRA-A matrix with random orthogonal weights and keeping it frozen during training, following the core idea of both.

\begin{table*}[t]
\centering
\resizebox{\textwidth}{!}{
\begin{tabular}{cccc|ccc|cccccc}
    \toprule
    \multirow{2}{*}{\textbf{Method}} & \multicolumn{3}{c|}{\textbf{4 concepts}} & \multicolumn{3}{c|}{\textbf{3 concepts}} & \multicolumn{3}{c}{\textbf{2 concepts}} & \multicolumn{3}{||c}{\textbf{Average}} \\
    ~ & CLIP-T\textuparrow & \makebox[0.04\textwidth]{CLIP-I\textuparrow} & DINO\textuparrow & CLIP-T\textuparrow & \makebox[0.04\textwidth]{CLIP-I\textuparrow} & DINO\textuparrow & CLIP-T\textuparrow & \makebox[0.04\textwidth]{CLIP-I\textuparrow} & DINO\textuparrow & \multicolumn{1}{||c}{CLIP-T\textuparrow} & \makebox[0.04\textwidth]{CLIP-I\textuparrow} & DINO\textuparrow \\
    \midrule
    Custom Diffusion   & .206 & .554 & .175 & .184 & .508 & .153 & .177 & 506 & .136 & \multicolumn{1}{||c}{.189} & .523 & .155 \\
    DB-LoRA & \textbf{.301} & .585 & .253 & \textbf{.294} & .604 & .284 & .291 & .621 & .294 & \multicolumn{1}{||c}{\textbf{.295}} & .603 & .277 \\
    Mix-of-Show  & \underline{.279} & \underline{.630} & \underline{.308} & .278 & \underline{.656} & \underline{.347} & .278 & \underline{.696} & \underline{.401} & \multicolumn{1}{||c}{.278} & \underline{.661} & \underline{.352} \\
    ROB  & .274 & .628 & .302 & .282 & .653 & \underline{.347} & \textbf{.296} & .675 & 378 & \multicolumn{1}{||c}{\underline{.284}} & .652 & .342 \\
    TARA (ours)  & .273 & \textbf{.633} & \textbf{.335} & \underline{.286} & \textbf{.673} & \textbf{.386} & \underline{.293} & \textbf{.711} & \textbf{.431} & \multicolumn{1}{||c}{\underline{.284}} & \textbf{.672} & \textbf{.384} \\
    \bottomrule
\end{tabular}
}
\caption{Quantitative evaluation of customized multi-concept generation. TARA achieves the best performance across all multi-concept settings on CLIP-I and DINO, significantly outperforming the DB-LoRA baseline and other popular methods. The scores are computed per concept and averaged to assess identity preservation and prompt alignment quality.}
\vspace{-1.5em}
\label{tab:multiconcept}
\end{table*}

\paragraph{Experimental Setup and Metrics.}
We conduct both quantitative and qualitative comparisons with the recent methods. For quantitative evaluation, we assess both single- and multi-concept generation using CLIP~\cite{radford2021learning} and DINO~\cite{caron2021emerging} features. Specifically, CLIP-T refers to the similarity between the generated image and its corresponding text prompt in the CLIP text-image embedding space, evaluating prompt alignment. CLIP-I denotes the cosine similarity between generated and reference images in the CLIP image embedding space, measuring identity preservation. DINO score is computed as the cosine similarity between generated and reference images using features from the pretrained DINO-ViT model, offering an additional view of visual similarity. For multi-concept generation, we report the average similarity across all involved concepts. To ensure robust evaluation, we generate 10 images for each prompt across all subjects using the prompt templates from the DreamBooth dataset. We also perform an ablation study under the multi-concept generation setting to evaluate the effectiveness of TFM and TAL.

\subsection{Quantitative Comparison}
\paragraph{Single Concept Comparison.}
Table~\ref{tab:single_concept} presents the quantitative results for single-concept generation before and after merging another LoRA module. The merging is used to test how the original concept changes after another concept (LoRA) is added. Our method achieves performance comparable to or better than other approaches in the single-LoRA setting, particularly in CLIP-I and DINO scores. More importantly, TARA exhibits the smallest performance drop after merging, with only a 0.004 drop in DINO on SD V1.5, while other methods often suffer significant degradation or require additional fusion time. These results highlight TARA’s robustness and efficiency in multi-LoRA composition, enabling near-instantaneous merging without sacrificing single-concept fidelity.
\begin{figure*}[t]
\centering
\includegraphics[width=0.9\textwidth]{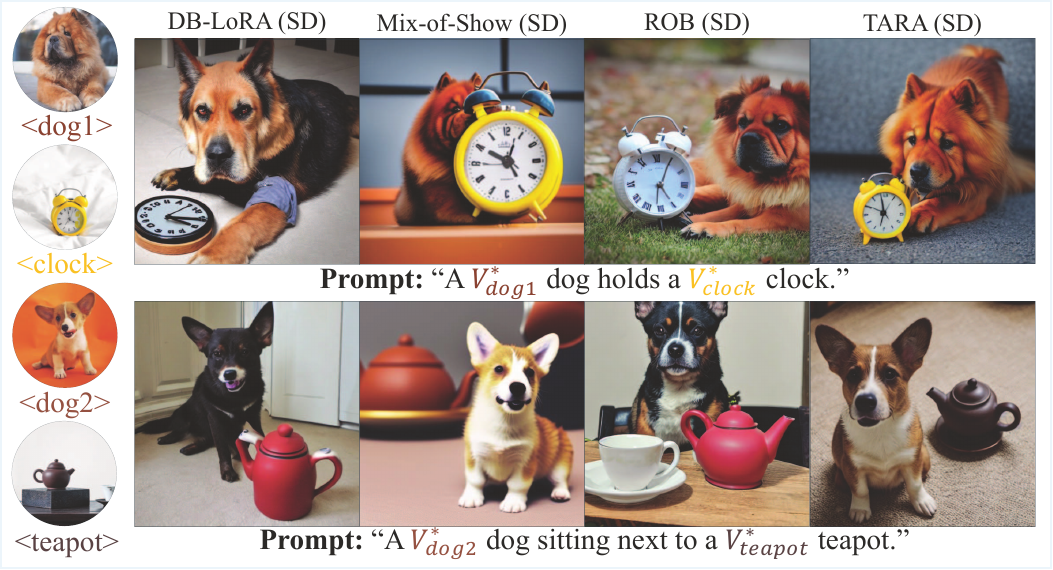} 
\vspace{-1.0em}
\caption{Qualitative comparisons between different LoRA-based methods. All images are generated using only text prompts, without additional training or external conditions. Compared to other methods, our approach better preserves the visual identity of each concept in the composed image.}
\label{qualitataive}
\end{figure*}
\begin{table}[t]
\centering
\resizebox{\columnwidth}{!}{
\begin{tabular}{c|lccc}
\toprule
 & \textbf{Component} & \textbf{CLIP-T} & \textbf{CLIP-I} & \textbf{DINO} \\
\midrule
(a) & LoRA & \textbf{.295} & .603 & .277  \\
(b) & + TFM & \underline{.284} & \underline{.662} & \underline{.374} \\
(c) & + TAL & .266 & .652 & .342 \\
(d) & TARA & \underline{.284} & \textbf{.672} & \textbf{.384} \\
\bottomrule
\end{tabular}
}
\caption{Ablation study of TARA in multi-concept generation. Both Token Focus Masking (TFM) and Token Alignment Loss (TAL) effectively improve identity preservation (CLIP-I and DINO) during multi-concept generation. Combining both leads to the best overall performance.}
\vspace{-1.5em}
\label{tab:ablations}
\end{table}

\paragraph{Multi-Concept Comparison.}
Table~\ref{tab:multiconcept} presents the quantitative results for multi-concept generation using the SD V1.5 backbone. TARA consistently achieves the highest scores across the identity preservation metrics (CLIP-I and DINO). For example, it achieves a CLIP-I score of 0.707 and a DINO score of 0.426 in the 2-concept setting, outperforming DB-LoRA (0.621 / 0.294) by large margins. Even under the challenging 4-concept setting, TARA maintains strong identity preservation (CLIP-I: 0.631, DINO: 0.335), while DB-LoRA drops significantly. Meanwhile, TARA’s prompt alignment remains comparable to other methods, indicating that prompt following capability is not compromised.

Importantly, Mix-of-Show not only fine-tunes the text encoder and text embeddings, but also requires additional training for fusion mechanisms. In contrast, TARA achieves these improvements without any extra training. These results demonstrate that TARA enables robust and efficient multi-concept composition with superior identity preservation and competitive prompt alignment.

\subsection{Qualitative Comparisons}
We compare our method with other LoRA-based approaches on multi-concept personalized generation. As shown in Figure~\ref{qualitataive}, our method achieves high-fidelity composition without requiring additional training or external conditions. Given only a user-provided prompt, our approach successfully integrates multiple concepts into a single coherent image, consistently outperforming other baselines in identity preservation and prompt adherence.

\subsection{Ablation Study}
To understand the contribution of each component in TARA, we conduct an ablation study under the multi-concept generation setting, as shown in Table~\ref{tab:ablations}. Starting from the standard LoRA baseline (a), we observe that adding Token Focus Masking (TFM) alone (b) significantly boosts identity preservation, with CLIP-I increasing from 0.603 to 0.662 and DINO from 0.277 to 0.374, while maintaining good prompt alignment (CLIP-T: 0.284). Applying Token Alignment Loss (TAL) alone (c) also improves CLIP-I (0.652) and DINO (0.342), though at the cost of a larger drop in CLIP-T. When both TFM and TAL are combined in the full TARA model (d), we achieve the best overall results: identity scores further improve (CLIP-I: 0.603 $\rightarrow$ 0.672, DINO:  0.277 $\rightarrow$ 0.384) while maintaining the same level of prompt alignment as TFM alone (CLIP-T: 0.284). These results confirm that TFM and TAL are complementary. TFM effectively identifies relevant tokens to focus on, and TAL strengthens spatial alignment. Together, they lead to more robust multi-concept generation.
\section{Conclusions}
We propose Token-Aware LoRA (TARA) for high-fidelity multi-concept personalization in diffusion models. Unlike DreamBooth LoRA, TARA applies token focus masking during the forward pass to constrain the effect of each injected LoRA module to its associated rare token, preventing interference caused by multiple LoRA modules acting on the same token. Additionally, we introduce a token alignment loss that guides the spatial attention of each rare token to align with its corresponding concept region, mitigating feature conflicts and improving generation quality. Together, these designs enable TARA to achieve strong identity preservation in multi-concept scenarios without requiring additional training or externally provided guidance.

{
    \small
    \bibliographystyle{ieeenat_fullname}
    \bibliography{main}
}

\end{document}